\documentclass[10pt,twocolumn,letterpaper]{article}

\usepackage[pagenumbers]{cvpr} 

\usepackage{graphicx}
\usepackage{amsmath}
\usepackage{amssymb}
\usepackage{booktabs}

\usepackage{multirow}
\usepackage{textcomp}
\usepackage{gensymb}
\usepackage{booktabs}
\usepackage{bbding}
\usepackage[noend]{algpseudocode}
\usepackage{algorithmicx,algorithm}

\usepackage[pagebackref,breaklinks,colorlinks]{hyperref}

\usepackage[capitalize]{cleveref}
\crefname{section}{Sec.}{Secs.}
\Crefname{section}{Section}{Sections}
\Crefname{table}{Table}{Tables}
\crefname{table}{Tab.}{Tabs.}

\begin{document}

\title{TAGPerson: A Target-Aware Generation Pipeline for Person Re-identification}

\author {
    Kai Chen\textsuperscript{\rm 1,3},
    Weihua Chen\textsuperscript{\rm 2},
    Tao He\textsuperscript{\rm 1,3},
    Rong Du\textsuperscript{\rm 2},
    Fan Wang\textsuperscript{\rm 2},
    Xiuyu Sun\textsuperscript{\rm 2},
    Yuchen Guo\textsuperscript{\rm 3},
    Guiguang Ding\textsuperscript{\rm 1,3} \\
    
    \textsuperscript{\rm 1} Tsinghua University, \textsuperscript{\rm 2} Alibaba Group \\
    \textsuperscript{\rm 3} Beijing National Research Center for Information Science and Technology (BNRist) \\
    \{chenkai2010.9, kevin.92.he, yuchen.w.guo\}@gmail.com, dinggg@tsinghua.edu.cn\\
    \{kugang.cwh, dr193339, fan.w,  xiuyu.sxy\}@alibaba-inc.com \\
     
}

\maketitle

\begin{abstract}
Nowadays, real data in person re-identification (ReID) task is facing privacy issues, e.g., the banned dataset DukeMTMC-ReID. Thus it becomes much harder to collect real data for ReID task. Meanwhile, the labor cost of labeling ReID data is still very high and further hinders the development of the ReID research. Therefore, many methods turn to generate synthetic images for ReID algorithms as alternatives instead of real images.
However, there is an inevitable domain gap between synthetic and real images. 
In previous methods, the generation process is based on virtual scenes, and their synthetic training data can not be changed according to different target real scenes automatically.
To handle this problem, we propose a novel Target-Aware Generation pipeline to produce synthetic person images, called TAGPerson. Specifically, it involves a parameterized rendering method, where the parameters are controllable and can be adjusted according to target scenes. In TAGPerson, we extract information from target scenes and use them to control our parameterized rendering process to generate target-aware synthetic images, which would hold a smaller gap to the real images in the target domain.
In our experiments, our target-aware synthetic images can achieve a much higher performance than the generalized synthetic images on MSMT17, i.e. 47.5\% vs. 40.9\% for rank-1 accuracy. We will release this toolkit\footnote{\noindent Code is available at \href{https://github.com/tagperson/tagperson-blender}{https://github.com/tagperson/tagperson-blender}} for the ReID community to generate synthetic images at any desired taste.
\end{abstract}

\section{Introduction}

\begin{figure}[ht]
\begin{center}
    \begin{subfigure}{1.0\linewidth}
        \includegraphics[width=1.0\linewidth]{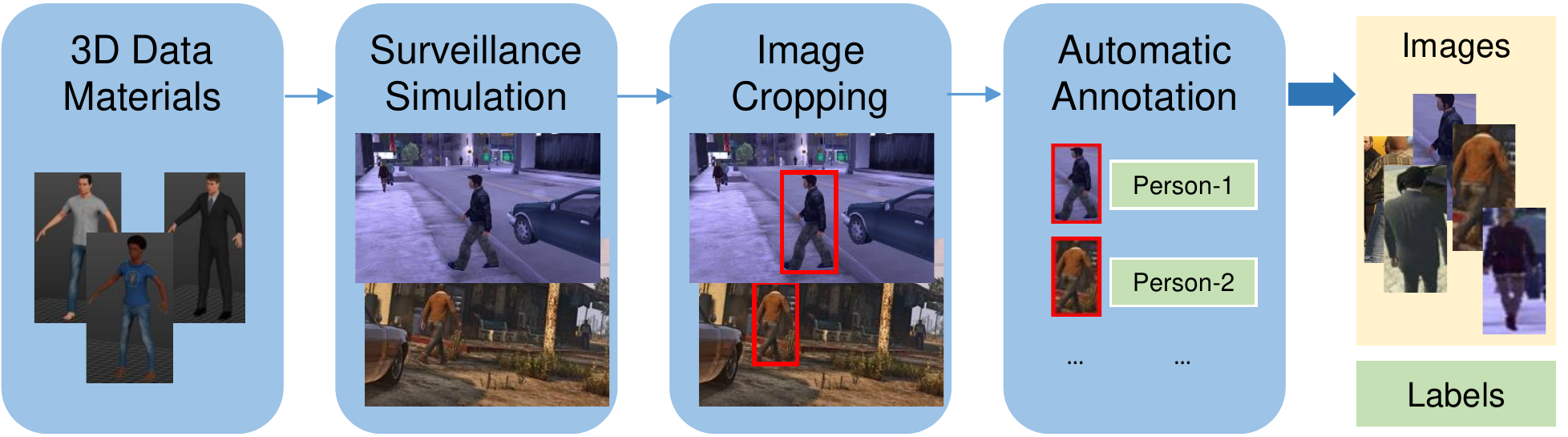}
    \vspace{-2mm}
    \caption{Synthetic datasets that based on scene simulation}
    \end{subfigure}
    \hfill
    \vspace{2mm}
    \begin{subfigure}{1.0\linewidth}
        \includegraphics[width=1.0\linewidth]{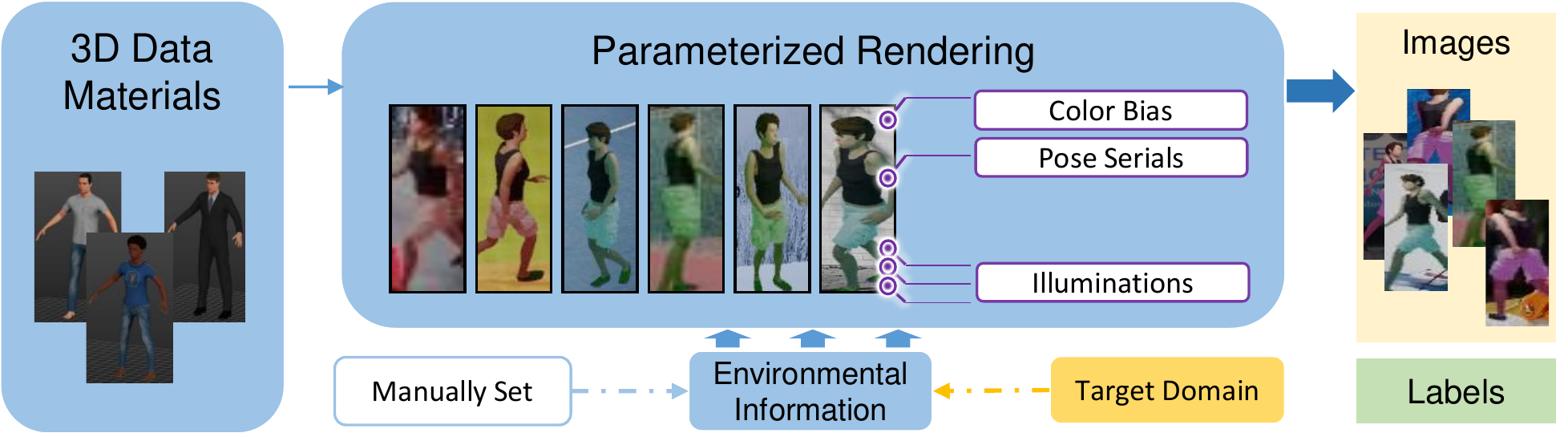}
    \vspace{-2mm}
    \caption{TAGPerson pipeline}
    \end{subfigure}
    \hfill
\end{center}
\vspace{-2mm}
\caption{Comparison between the building process of current synthetic datasets and proposed TAGPerson. (a) A popular workflow of current synthetic datasets which are based on scene simulation. (b) Our proposed TAGPerson is based on parameterized rendering, and it can integrate information from the target domain during the rendering process.}
\label{fig:comparison_of_pipelines}
\vspace{-2mm}
\end{figure}

Person re-identification (ReID) technology is expected to determine if two persons across different views have the same identity. It plays an important role in intelligent applications of surveillance scenarios and draws a lot of attention in the computer vision area \cite{ye2021deep}. With the development of deep learning, several large-scale annotated pedestrian datasets \cite{zheng2015scalable, wei2018person} have been produced, and the fully supervised methods for person ReID have achieved great success\cite{luo2019bag, sun2018beyond}. However, with the increasing awareness of privacy protection, images containing human biology information are often along with ethical issues \cite{wu2021decentralised}. The identity labels are also difficult to label between cross-scene and cross-camera views, making the labeling cost very high \cite{wang2020surpassing}. 

Several synthetic pedestrian datasets have been proposed to alleviate these problems \cite{barbosa2018looking,bak2018domain,sun2019dissecting,wang2020surpassing,zhang2021unrealperson}. Synthetic datasets benefit from data diversification at low cost and the automatic generation of annotated labels.
Recent synthetic datasets \cite{sun2019dissecting,wang2020surpassing,zhang2021unrealperson} are based on scene simulation. They construct virtual scenes and capture screenshots of the pedestrians who are walking around there. 
However, they suffer from the domain gap between virtual and real images. 
The scene-based rendering process is a snapshot sampling of the virtual scenes, and the synthesized images depend on the manually set scenes. 
The synthetic training data can not be changed according to different target real scenes automatically.
To make up for this shortcoming, this paper proposes TAGPerson, a Target-Aware Generation (TAG) pipeline to generate the auto-labeling synthetic ReID datasets. Our method can integrate the target domain information during the rendering process to narrow the gap between synthetic data and real data.

TAGPerson is composed of three stages. 3D data materials are firstly prepared to provide the basic person characters. Next, images are rendered under parameterized control by manual setting or the target domain. It is achieved by manual setting from prior knowledge or estimated distribution statistics of the target domain. The estimation models are trained to extract information from the target domain. The rendered images together with their labels serve as the training data to train the ReID model.
The difference between our workflow and the previous synthetic pipeline can be viewed in \cref{fig:comparison_of_pipelines}. 

Our method has advantages in three aspects. First, we integrate the target domain information during the rendering process. The rendering options can be constrained by the target domain, making the rendered images hold a smaller gap to the real images in the target domain.
Existing pipelines are oriented towards general scenarios. They do not take into account the utilization of possible available information about the target domain.
Second, our parameterized rendering decouples different factors about the composition of the image, which reveals the most important environmental factors that affect the performance of the ReID model. 
Last, we can simulate more variables quantificationally like observation angle of view and illumination conditions. This feature facilitates the rendering process to deal with some extreme scenarios.

Based on this procedure described above, TAGPerson has opened up a new path to achieve integrating information from the target domain. 
Our main contributions can be summarized as follows:
\begin{itemize}
    \item We propose a novel target-aware person ReID dataset construction process. The synthetic data are auto-labeled and the images can be rendered under parameterized control, under the guidance of the target domain information. Based on the synthesized data, many person Re-ID tasks can be resolved without using the real datasets.
    \item We come up with a simple yet effective solution to extract the information from the target domain, and use it to guide the rendering process. Images of persons are rendered in a target-aware manner, which significantly improves the generalization ability of the person ReID models in real scenarios.
    \item Experiments are conducted to explore the effects of key factors during the rendering process. The gap of performance between the usage of synthetic and real images can be narrowed, which enhances the availability of ReID models in the limited scenario where there exist data restriction and privacy issues. 
\end{itemize}

\section{Related Work}

\subsection{Person ReID Tasks}
Person ReID task is welcomed since it can be used to solve realistic problems in surveillance scenarios. Metrics-based learning methods are proposed to measure the similarity of the query probe and the images in the gallery \cite{hermans2017defense,varior2016gated,sun2020circle}. Supervised person ReID has achieved great success in recent years \cite{luo2019bag,wang2018learning,li2014deepreid,porrello2020robust}, but the models usually get degraded performance when deployed into another scenario. Unsupervised domain adaptation methods \cite{li2020joint,zheng2021group,bai2021unsupervised,xuan2021intra,yang2021joint,jin2020global} are studied by researchers to improve the effectiveness on target domain. These methods can be split into two categories. Some of them use the basic model trained from the source data to assign pseudo labels for the target images, and then iteratively update the model and the label assignment \cite{ge2019mutual,NEURIPS2020_821fa74b}. Other solutions use GAN-based methods to transfer the source images into styles of target domain \cite{CycleGAN2017,deng2018image,zhong2018camera,jin2020style}, which alleviates the gap between the two domains. 
The above methods rely heavily on labeled or unlabeled data, which might not be satisfied under the circumstances of data privacy protection. In recent years, researchers have resorted to synthetic methods to generate pedestrian datasets in the absence of real data.

\begin{figure*}[ht]
\begin{center}
    \includegraphics[width=1.0\linewidth]{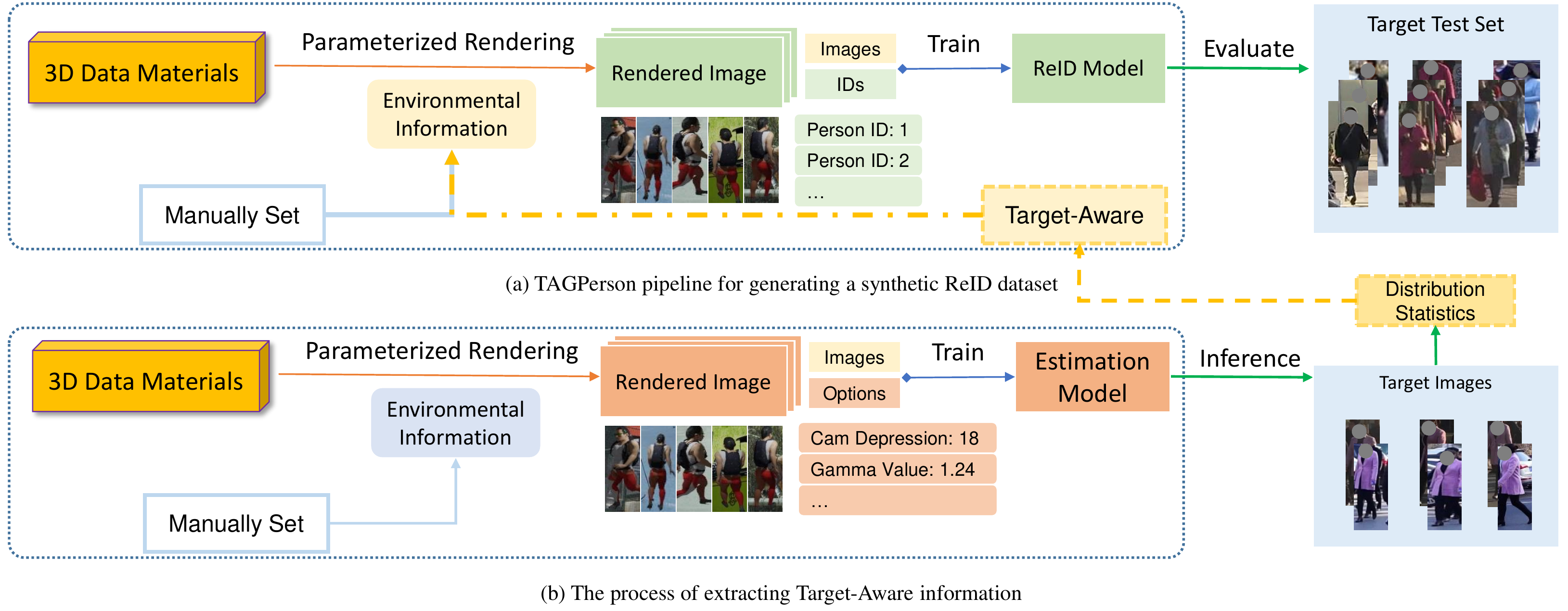}
\end{center}
\vspace{-2mm}
\caption{The illustration of the TAGPerson pipeline. Firstly, 3D data materials are prepared. Then the images are rendered under parameterized control. The rendered images together with their labels are used to train the ReID model or estimation models. (a) To generate a synthetic ReID dataset, the environmental information for parameterized rendering is set manually or from the target domain information. (b) The rendered images with their rendering option labels are used to train estimation models, which are used to extract information from the target images. The distribution statistics are passed to guide the rendering process and make it target-aware.}
\label{fig:dataset_construction_procedure}
\end{figure*}

\subsection{Synthetic Datasets}
In consideration of the problems about data privacy and ethical issues, many researchers resort to constructing synthetic pedestrian datasets for ReID task. SOMAset \cite{Bengio+chapter2007} and SyRI \cite{bak2018domain} open the door to use synthetic datasets in deep learning ReID models. SyRI\cite{bak2018domain} explore the effect of illumination for persons by using GAN-based methods to synthesize the images. PersonX \cite{sun2019dissecting} is a large-scale person dataset constructed by Unity3D, which explores the influence of pedestrian rotation angle on ReID accuracy. RandPerson \cite{wang2020surpassing} proposes a texture generation method to produce masses of virtual persons and establishes a set of customized environments to simulate the surveillance scenes. Recently, UnrealPerson \cite{zhang2021unrealperson} analyzes the strategies during 3D human data generation. It proves that hard samples in training data are important to improve performance. UnrealPerson uses Unreal Engine 4 and UnrealCV \cite{qiu2016unrealcv} to simulate real-world scenes, and achieves excellent results in several kinds of person ReID tasks.

Previous scenario-oriented synthetic datasets have domain gaps between synthetic and real images. 
To solve this problem, we construct our TAGPerson dataset in a target-aware manner by integrating the target domain information during the rendering process.
All identity-unrelated factors are constructed for a single target pedestrian, including different character poses, background images, camera perspectives, light conditions, etc. These rendering options can be under the guidance of the target domain information.

\section{Our Method}

\begin{table*}[t]
\caption{Detailed comparisons of synthesized datasets. TAGP-Base means the rendering options are set manually. TAGP-TA means some of the rendering options are controlled by the target domain. Note that we do not create the virtual scene and camera network. By replacing background images and adding color bias, the count of our cameras can be regarded as infinite. The rank-1 accuracy on Market-1501 and MSMT17 datasets is the direct transfer performance of ReID models trained on the synthesized datasets.}
\label{table:dataset_comparison}
\vspace{-3mm}
\begin{center}
\begin{tabular}{|l|c|c|c|c|c|c|c|}
    \hline
    Datasets & \#Identities & \#Cameras & \#BBoxes & \shortstack{Parameterized \\Rendering} & Scalabel & \shortstack{Rank-1 on\\ Market} & \shortstack{Rank-1 on \\ MSMT} \\
    \hline
    \hline 
    SyRI \cite{bak2018domain} & 100 & - & 56,000 & \XSolidBrush & \XSolidBrush & 48.5\% & 21.8\%\\
    PersonX \cite{sun2019dissecting} & 1266 & 6 & 273,456 & \XSolidBrush & \Checkmark & 58.7\% & 22.2\% \\
    RandPerson \cite{wang2020surpassing} & 8000 & 19 & 228,655 & \XSolidBrush & \Checkmark & 64.7\% & 20.0\% \\
    UnrealPerson \cite{zhang2021unrealperson} & 3000 & 34 & 120,000 & \XSolidBrush & \Checkmark & 79.0\% & 38.5\% \\
    \hline
    TAGP-Base & 2954 & infinite & 71,580 & Manual & \Checkmark & 79.9\% & 40.9\% \\
    TAGP-TA & 2954 & infinite & 71,580 & Target Aware & \Checkmark & 81.6\% & 47.5\% \\
    \hline
\end{tabular}
\end{center}
\vspace{-3mm}
\end{table*}

\subsection{TAGPerson Pipeline}

Existing synthetic person ReID datasets usually build virtual scenes and then capture person images there.
Different from previous synthetic methods, we use a parameterized rendering process to directly generate images under specific rendering options. We avoid the process of scene simulation and data sampling, in order to obtain more control over the factors which dominate the quality of the person ReID dataset. 
Suppose the training dataset $S_{train} = \{\cup_{k=1}^K (I_k, y_k)\}$ consists of image $I_k$ and its identity label $y_k$. From the parameterized perspective, the content of image data $I_k$ can be decoupled into two parts: identity-related information and identity-unrelated information:
\begin{equation}
    I_k = (P_k, \sum_{j=1}^J {F_{o_j}})
\end{equation}
where the $P_k$ represents identity-related information for person $k$ and $F{o_j}$ represents the identity-unrelated environmental factor which is caused by the rendering option $o_j$. The rendering option $o_j$ can be background, image resolution, illumination, pose, camera parameters, etc. That is what we can control during the rendering process.

Our proposed TAGPerson can be divided into three stages: prepare the 3D data materials, render the images, and train the ReID model. The overall pipeline can be viewed in \cref{fig:dataset_construction_procedure}.
During the construction, all the identity-related contents are determined by the 3D data, and other identity-unrelated contents are generated by the parameterized rendering. The environmental information can be set manually or from the target domain information. The rendered images with their labels are used to train the ReID model or estimation models.

As a prerequisite, the 3D data of humans are the fundamental content to build a person ReID dataset. They directly make up the appearances of humans. 
We control many factors to generate the 3D data for one person, including its skin, face, height, obesity, muscle, etc. To distinguish one person from another, we dispatch different clothes, dresses, hairstyles, shoes, and optional accessories for each person. Next, the parameterized rendering process will be introduced in detail.

\subsection{Parameterized Rendering}
\begin{figure}[ht]
\begin{center}
    \begin{subfigure}{0.48\linewidth}
        \includegraphics[width=1.0\linewidth]{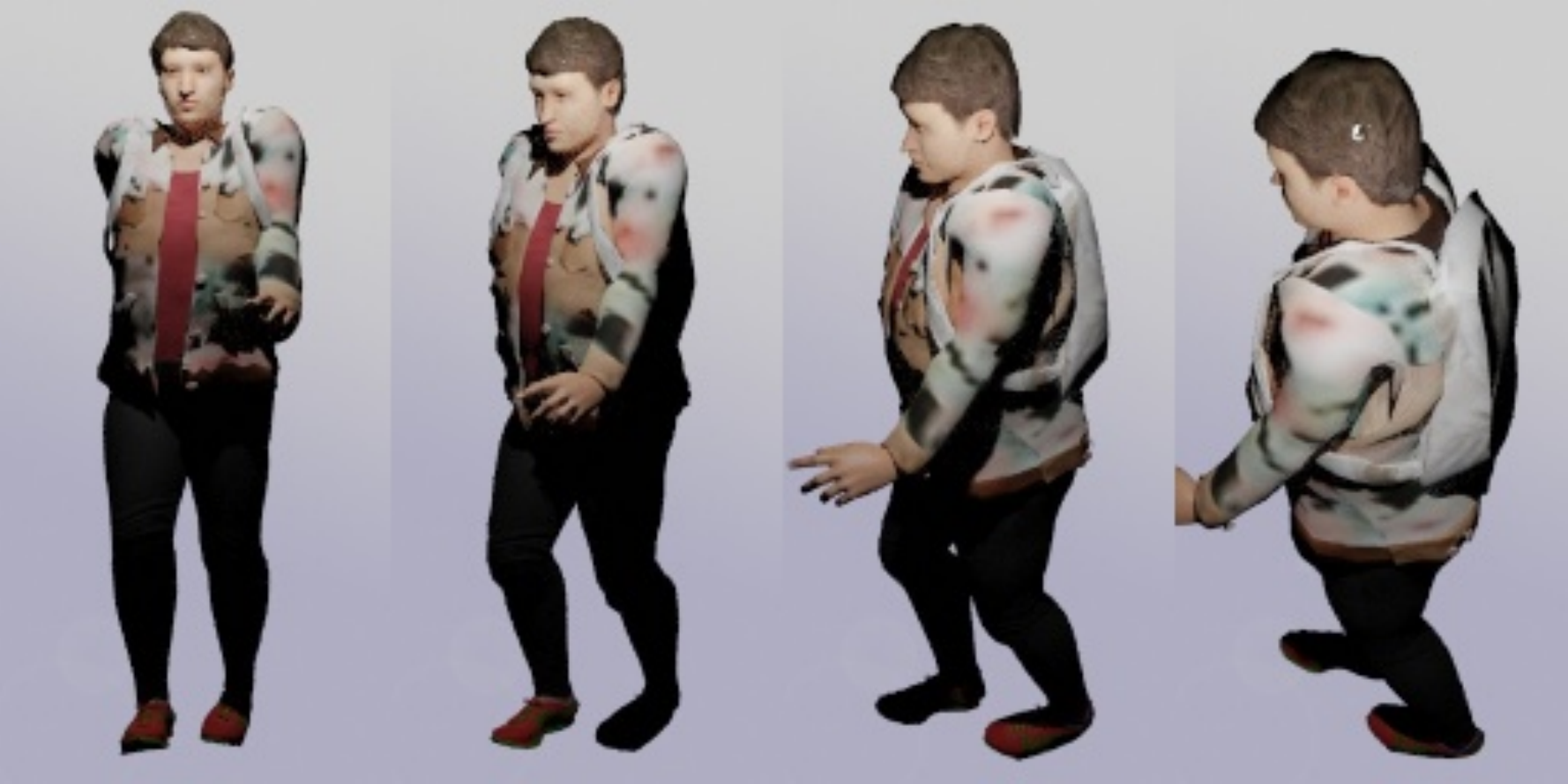}
    \caption{Camera Azimuth}
    \end{subfigure}
    \hfill
    \begin{subfigure}{0.48\linewidth}
        \includegraphics[width=1.0\linewidth]{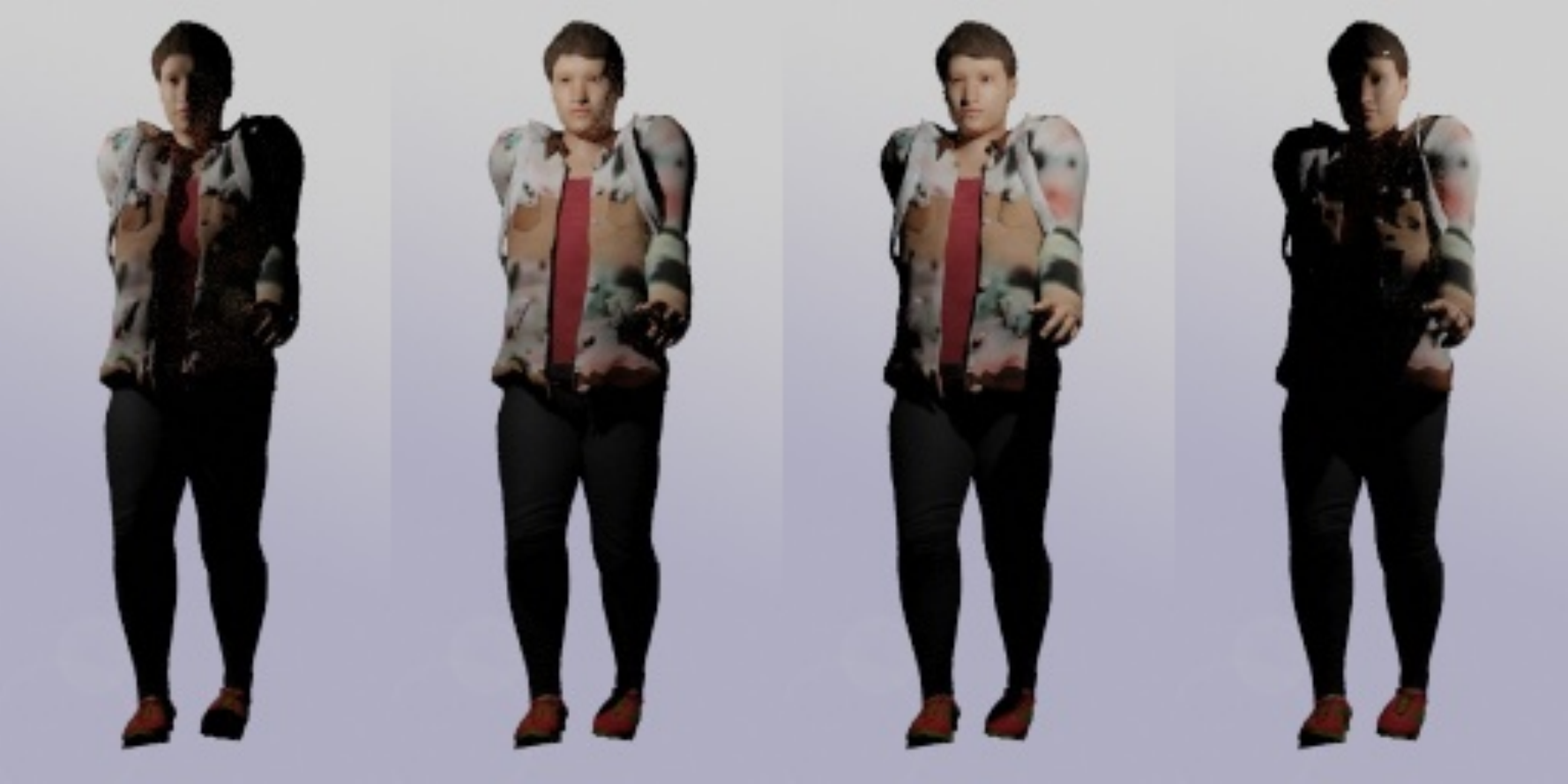}
    \caption{Illumination}
    \end{subfigure}
    \hfill
    \begin{subfigure}{0.48\linewidth}
        \includegraphics[width=1.0\linewidth]{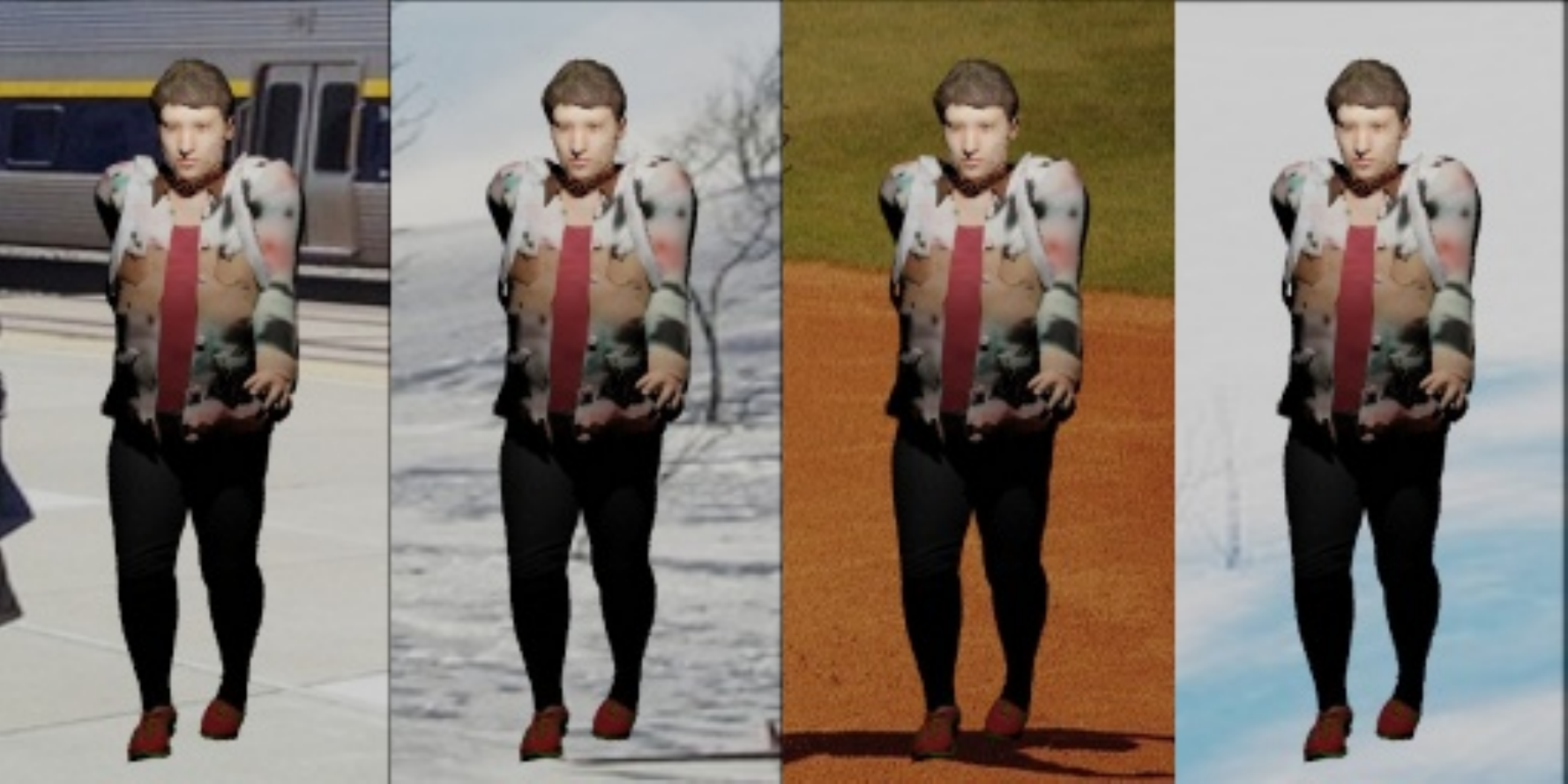}
    \caption{Background}
    \end{subfigure}
    \hfill
    \begin{subfigure}{0.48\linewidth}
        \includegraphics[width=1.0\linewidth]{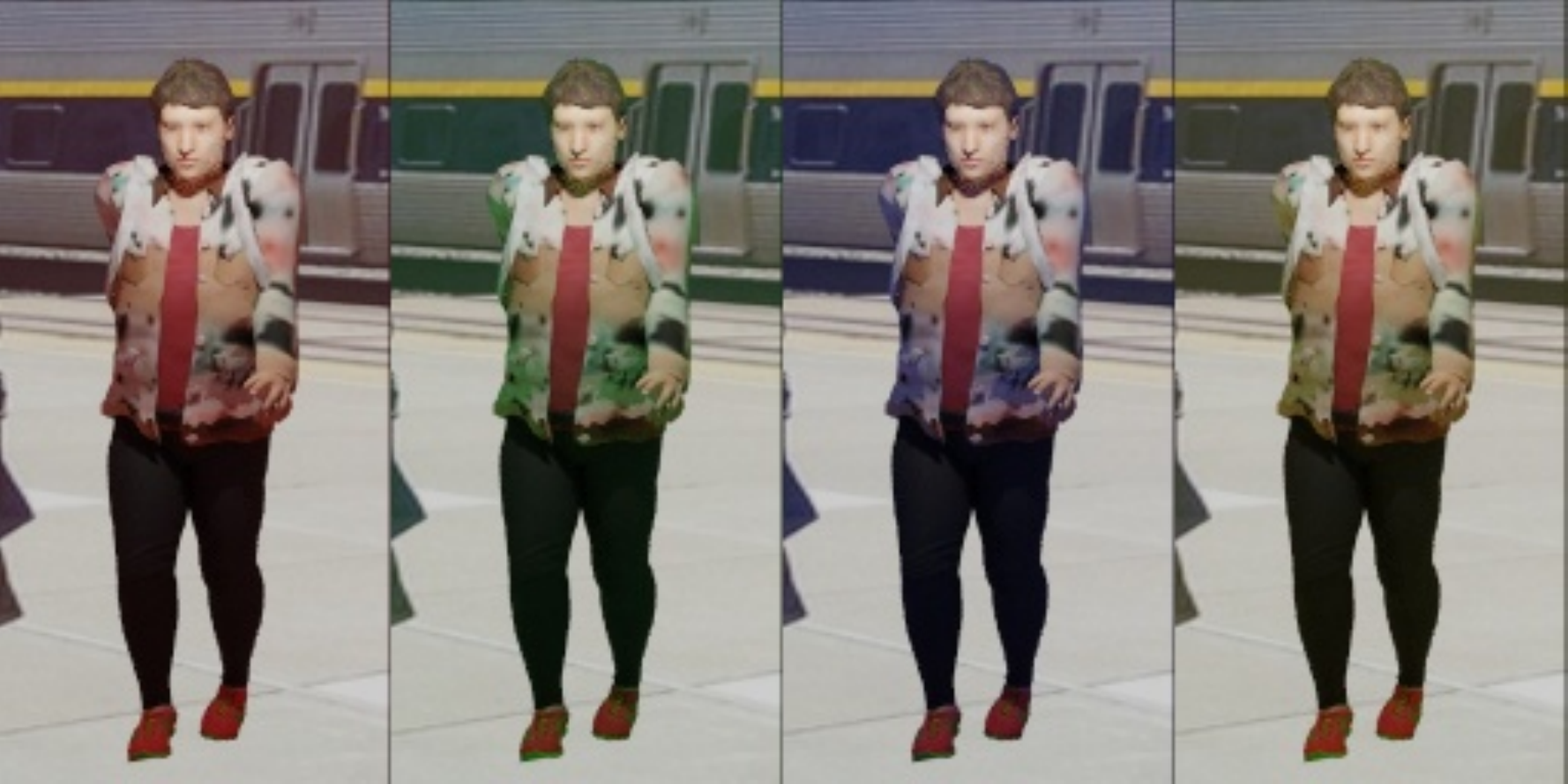}
    \caption{Color Bias}
    \end{subfigure}
\end{center}
\vspace{-5mm}
\caption{Visualizations of the effects of some rendering options. (a) Different camera azimuths. (b) Different light conditions. (c) Changing background images. (d) Adding color bias. All the rendering options can be parametrically controlled, and a pedestrian individual can be rendered from all aspects.}
\vspace{-5mm}
\label{fig:render_result}
\end{figure}

In previous works about synthetic datasets, the environmental information is determined by the virtual scenes. The rendering options are rarely mentioned. 
For the same 3D human data, different rendering options can produce different images.
We dig into some rendering options that may affect the performance of the final model. 
The example rendered images can be seen in Fig.~\ref{fig:render_result}. 

\noindent
{\bf Rig Pose.}
We increase the variety of poses by introducing the change to the bones rig. Motion capture is the process of recording the movement of objects, and CMU Graphics Lab has published a free Motion Capture Database \cite{gross2001cmu}. By applying the \verb+bvh+ files, we can make the person change its pose to a specific one, which can be an arbitrary moment of walking, standing, or others.

\noindent
{\bf Camera Parameters.}
Existing methods may use a detection model or segmentation annotations to crop the pedestrian bounding boxes. 
We adopt another way by putting the target person as the anchor and setting different camera positions and orientations to capture images. 

\noindent
{\bf Illumination.}
Illumination conditions affect the appearance of the person to a great extent. 
Previous works preset the lights in the virtual scenes, where each person's illumination condition is determined by its position. Instead, we change the position and intensity of the light source to produce controllable diverse illumination conditions from all angles for the target person.

\noindent
{\bf Image Resolution.}
The distribution of image resolution is easily overlooked in previous datasets. If images have been resized to a fixed size, the intuitive representation is the degree of blurring of the images. In the real world, how far the person is from the camera directly determines the resolutions of the cropped images. 
From the experiments, we found that we can improve the robustness of the ReID models by setting diversified image resolutions properly.

\noindent
{\bf Background.}
The background is the important factor during the process of person ReID, which always contains a certain domain-specific message of the current dataset. Without creating a virtual scene, a trivial solution is to use images of diverse scenes as the background image to make the generated dataset robust to different complex scenarios.
In our case, we use images from COCO \cite{lin2014microsoft} dataset as the background images. Each image is appropriately cropped according to the annotation of the person instance. 

\noindent
{\bf Color Bias.}
To make features distinguishable from the global perspective, many methods have attempted to eliminate the existing camera bias in images. We resolve the problem of camera style differences by simulating images with color bias quantitatively. We use a simple yet efficient strategy to add specific color bias to images according to the camera label currently assigned.

\subsection{Target-Aware Generation}
We intend to integrate the target domain information during the rendering process, towards making the rendered images more inclined to the style of the specific target domain.
We extract the target-aware information from the target domain, and then apply it in reverse to control the rendering options. We use $O={o_1, o_2, ..., o_J}$ to represent the rendering options. For example, $o_j$ can be the camera depression angle towards the person. We use $D^t(O)$ to describe the distribution of factors $O$ on the target domain, and $D^s(O)$ as the one on the current synthetic domain. The Wasserstein Distance \cite{vallender1974calculation} can be used to measure the discrepancy between $D^s(O)$ and $D^t(O)$ as $W(D^s(O), D^t(O))$, and our goal is to minimize the discrepancy.
\begin{equation}
    \mathop{min}_{D^s(O)} W(D^s(O), D^t(O))
\end{equation}

We design a mechanism to automatically fetch the target-specific information. First, by manually setting the rendering options, we generate masses of images with their rendering option labels. These data are used to train an estimation model $M_{r_j}$ which can be used to inference the value of rendering option $o_j$ for an input image $I^t_i$: 
\begin{equation}
    \tilde{o_{ij}} = M_{r_j}(I^t_{i})
\end{equation}
Here $I^t_i$ means the $i$th image of the target domain and $\tilde{o_{ij}}$ is the output value. We use all the available images of the target domain as the input fed to the model. The values of the inference output are collected to form distribution statistics. 
\begin{equation}
    D^t(o_j) \sim \sum_{i=1}^{N_t}{\tilde{o_{ij}}}
\end{equation}
\begin{equation}
    D^t(O) = \cup_{j=1}^{J}{D^t(o_j)}
\end{equation}
Here $N_t$ is the total number of collected target images, and $D^t(o_j)$ represents the distribution of factor $o_j$ on the target domain.  $J$ is the number of rendering options. We fetch the distribution statistics for all $J$ necessary rendering options to generate the distribution statistics $D^t(O)$. It is used as the target-aware information to guide the parameterized rendering process.

During the process of synthetic dataset generation, the rendering options are constrained by the above distribution $D^t(o)$ to fit the style of the target domain. We use a simple yet effective Monte Carlo \cite{hammersley2013monte} sampling method to ensure the limitation of $D^s(O)$. To be precise, the values of rendering options are randomly sampled from the distribution statistics $D^t(o)$ with equal probability. This strategy makes the images rendered in a target-aware manner, and the rendered images are closer to the target domain in terms of the distribution of corresponding environmental factors.
We assumed that these rendering options are less affected by the cross-domain problem compared to identification labels during the estimation model training, and the estimation error can be inessential. 

Note that our approach does not need to read the images of the target domain directly, and the images of the target domain are not involved in the training process. The information about the target domain is extracted and inducted from the images, which is less sensitive and makes it possible to avoid data privacy issues. In the practical application scenarios, the extraction methods can be delivered to the data owner, and they give back the statistics extracted from the data rather than the whole original data itself.

\section{Experiments}

In this section, we introduce our implementation details and the experimental results. 
The TAGPerson dataset is generated by manually set or target-aware rendering options. The overview of generated TAGPerson dataset can be seen in \cref{table:dataset_comparison}. Some experiments are conducted to explore the importance of rendering options and demonstrate the effectiveness of our target-aware mechanism.

\subsection{Implementation Details}
During the data generation, we utilize the MakeHuman \cite{makehuman} Python API to generate thousands of 3D data of humans. We render images by Blender Python Library \cite{blender}. In the training stage, We use the Fastreid \cite{he2020fastreid}, a toolkit based on PyTorch \cite{paszke2019pytorch}, as the basic training framework. 
We use ResNet-50 \cite{he2016deep} structure as the backbone, which is pre-trained on ImageNet \cite{deng2009imagenet}.
We train the model based on the labeled synthetic images with cross-entropy loss and triplet loss, and then directly evaluate it on the target datasets. 
The input images are resized to $256 \times 128$.
We use SGD as the optimizer with a momentum of $0.9$ and weight decay of $0.0005$.
The initial learning rate is $0.05$, and it decays to $0.005$ and $0.0005$ after $30$ and $60$ epochs. The training stops at $80$ epochs.
We set the mini-batch size to $32$. 
In each mini-batch, $4$ identities are chosen and $8$ images of each identity are randomly sampled.
We choose ColorJitter and AugMix \cite{hendrycks2019augmix} as data augmentation.

\begin{table}[t]
\caption{Direct transfer performance of some real datasets and synthetic datasets.}
\vspace{-5mm}
\label{table:direct_trainsfer}
\begin{center}
\small
\begin{tabular}{|c|l|cc|cc|}
    \hline
    \multirow{2}{*}{Source} & \multirow{2}{*}{Training Data} & \multicolumn{2}{c|}{Market} & \multicolumn{2}{c|}{MSMT} \\
    \cline{3-6} 
         & & R1 & mAP & R1 & mAP  \\
    \hline
    \hline 
    \multirow{2}{*}{Real} & Market & 94.7 & 86.2 & 25.7 & 9.6 \\
     & MSMT17 & 74.4 & 45.4 & 74.7 & 50.5 \\
    \hline
    \multirow{4}{*}{Synthetic} & SyRI & 48.5 & 22.6 & 21.8 & 5.7 \\
    & PersonX & 58.7 & 32.7 & 22.2 & 7.9 \\
    & RandPerson & 64.7 & 39.3 & 20.0 & 6.8 \\
    & UnrealPerson & 79.0 & 54.3 & 38.5 & 15.3\\
    \hline
    \multirow{2}{*}{Synthetic} & TAGP-Base & 79.9 & 53.1 & 40.9 & 14.3 \\
    & TAGP-TA & \bfseries{81.6} & \bfseries{54.8} & \bfseries {47.5} & \bfseries {17.7} \\
    \hline
\end{tabular}
\vspace{-5mm}
\end{center}
\end{table}

\begin{table*}[ht]
\caption{Detailed comparisons of rendering options. The check symbol means that the rendering option in the column has proper multiple values, otherwise it is set to a default value. Various background images are necessary and different resolutions are beneficial to ReID performance. Multiple illumination conditions and diversified poses help to improve the effect. Different camera depression angles can promote the performance. Slightly adding color bias makes the model more robust.}
\vspace{-5mm}
\label{table:rendering_options}
\begin{center}
\begin{tabular}{|c|c|c|c|c|c|cc|cc|}
    \hline
    \multirow{2}{*}{Background} & \multirow{2}{*}{Resolution} & \multirow{2}{*}{Illumination} & \multirow{2}{*}{Pose} & \multirow{2}{*}{ \shortstack{Camera \\ Depression Angle}} & \multirow{2}{*}{\shortstack{Color \\ Bias}} & \multicolumn{2}{c|}{Market} & \multicolumn{2}{c|}{MSMT} \\
    \cline{7-10}
        & & & & & & R1 & mAP & R1 & mAP \\
    \hline
    \hline 
        -& -  & -   & -    &    - & -  & 22.1 & 8.6 & 3.9 & 1.1 \\
    \checkmark &   -   &   -  &    -  &   -   &   - & 65.8 & 38.5 & 29.1 & 9.2 \\
    \checkmark & \checkmark &   -  &   -   &  -    &  -  & 74.5 & 47.0 & 31.2 & 9.9 \\
    \checkmark & \checkmark & \checkmark &    -  &   -   &  -  & 75.2 & 47.1 & 34.0 & 11.2 \\
    \checkmark & \checkmark & \checkmark & \checkmark &  -    &   - & 76.1 & 49.4 & 35.0 & 12.0 \\
    \checkmark & \checkmark & \checkmark & \checkmark & \checkmark &  -  & 78.7 & 54.0 & 36.2 & 12.8 \\
    \checkmark & \checkmark & \checkmark & \checkmark & \checkmark & \checkmark & 79.9 & 53.1 & 40.9 & 14.3 \\
    \hline
\end{tabular}
\vspace{-5mm}
\end{center}
\end{table*}

\subsection{Direct Transfer Evaluation}
Direct transfer evaluation is the setting closest to the actual application scenario where the target domain is unavailable during training. To prove the validity and practicality, we apply direct transfer evaluation for the model trained by our proposed TAGPerson dataset on two real-world person ReID datasets. Market-1501 \cite{zheng2015scalable}, MSMT17 \cite{wei2018person} are used as the test sets. The direct transfer performance of TAGPerson and other datasets is compared in \cref{table:direct_trainsfer}.

For short we use Market and MSMT to represent Market-1501 and MSMT17. TAGP-Base represents that the rendering options are manually set and TAGP-TA represents that the rendering options are guided by the target domain information. From the table, we can see that without relying on the construction of multiple virtual scenes, TAGP-TA can achieve competitive performance compared to the state-of-the-art method in synthetic datasets. The performance on the Market dataset also surpasses the performance of the large-scale real-world dataset MSMT. Specifically, we boost the rank-1 accuracy and mAP on Market to $81.6\%$ and $54.8\%$, respectively. On the MSMT dataset, our TAGP-TA boosts the rank-1 accuracy and mAP to $47.5\%$ and $17.7\%$, surpassing all previous synthetic datasets.

\subsection{Ablation Study on Rendering Options}

Rendering options have significant impacts on the performance of the ReID model. We explore the effects of several rendering options and find the most important factors that contribute to good synthetic images. The results can be viewed in \cref{table:rendering_options}. 

The experimental results show that complex background information is the most important necessity. By using the background images cropped from the COCO \cite{li2020joint} dataset, the performance has a huge improvement for rank-1 accuracy from $3.9\%$ to $29.0\%$ on the MSMT dataset, compared to the case of using empty background.
Different sizes of the resolution also bring large improvements to the performance, especially on the Market dataset. The mAP on the Market dataset is boosted from $38.5\%$ to $47.2\%$ because there are many blurred images there.
Multiple illuminations have positive effects, especially on the MSMT dataset where some images are under extreme lighting conditions. 
Applying various poses to the person increases the diversity and improves the performance, which is reasonable. 
Multiple camera depression angles can deal with the situation when the heights of cameras are different, and they have positive effects when introduced. The improvement is not obvious since the range is set manually, without considering specific scenarios.
Adding the color bias factor boosts the rank-1 accuracy and mAP on the MSMT dataset over $4.7\%$ and $1.5\%$ respectively.

\subsection{Ablation Study on Target-Aware Rendering}

Camera and illumination are important factors in person ReID task \cite{zhong2018camera,loy2009multi,ma2019low}. 
However, from the above experiments, we find that adding multiple camera parameters and illumination conditions does not bring obvious improvement. 
We suppose that these factors are strongly correlated with the dataset. they differ in different datasets and it is hard to control the value range manually. 
That hinders the improvement of the performance on real-world datasets.

To solve this problem, we conduct experiments to demonstrate the effectiveness of our proposed target-aware generation for the rendering process. We adopt camera depression angle and gamma value as the representatives of target-aware information to control the rendering options. Synthetic datasets with different rendering options are generated. We compare the performance of ReID models trained based on them.

\begin{figure*}[ht]
\centering
  \begin{subfigure}{0.33\linewidth}
    \includegraphics[width=1.0\linewidth]{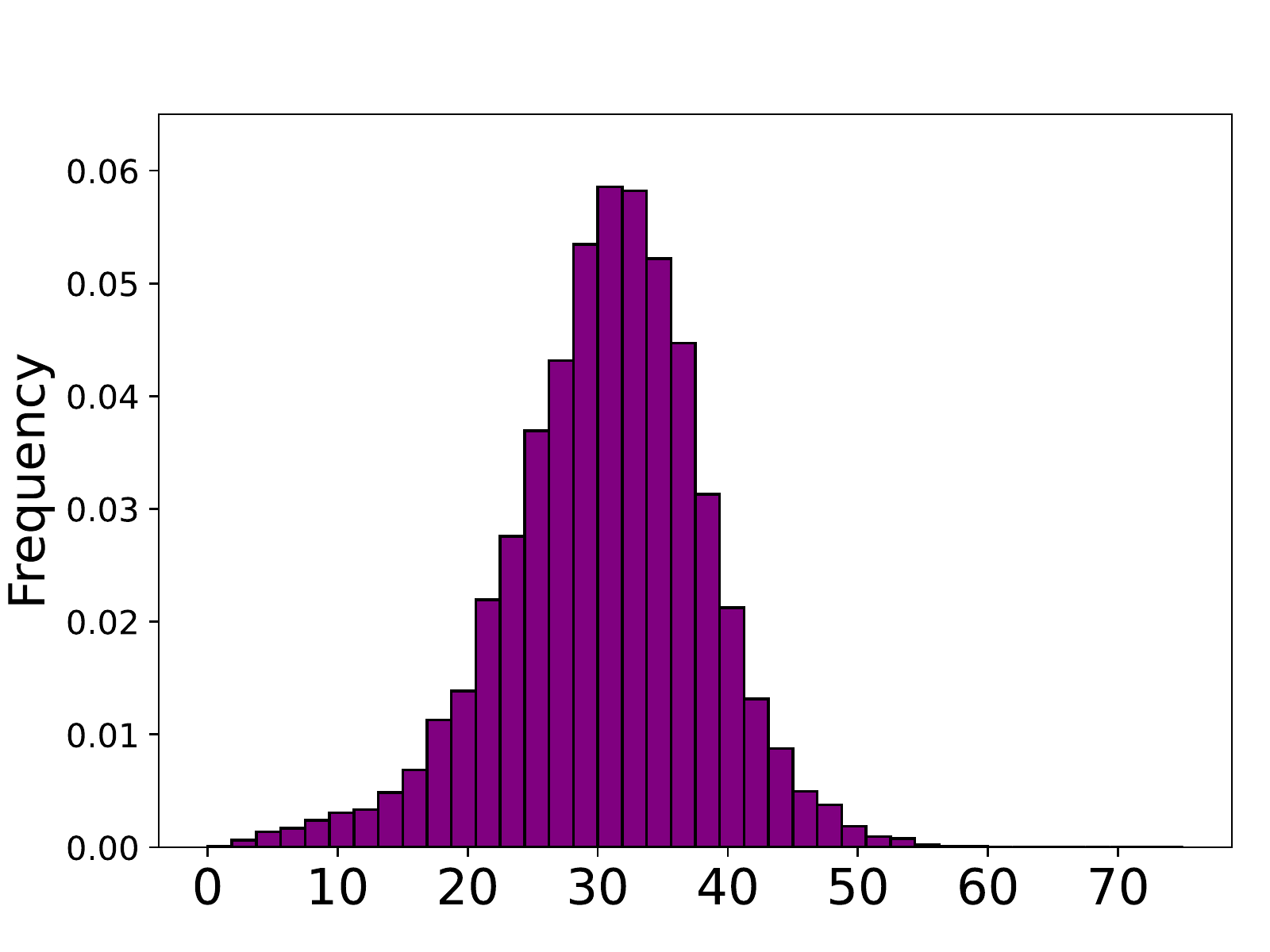}
    \caption{Market-1501}
    \label{fig:distribution_camera_depression_angle_1}
  \end{subfigure}
  \hfill
  \begin{subfigure}{0.33\linewidth}
    \includegraphics[width=1.0\linewidth]{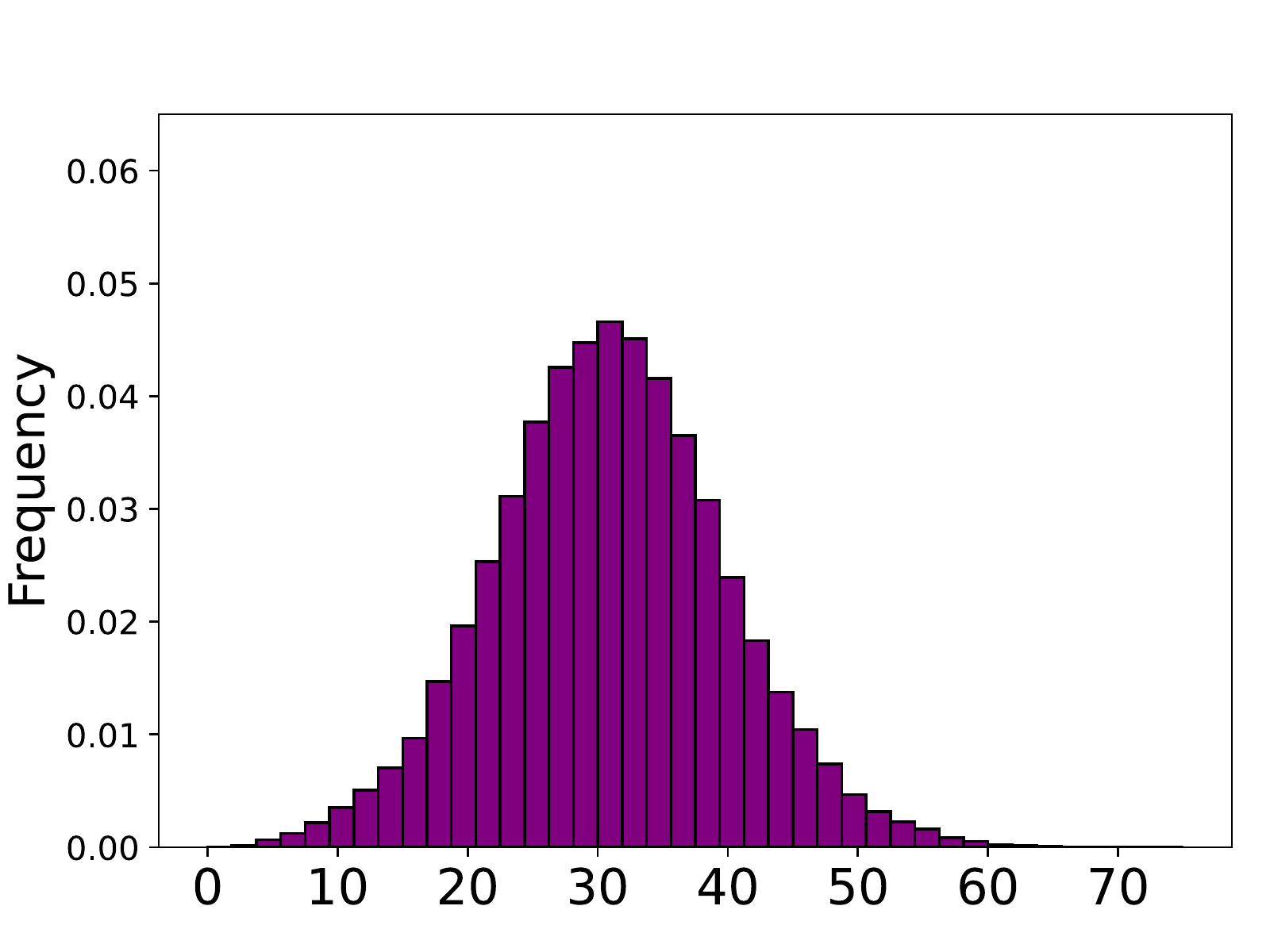}
    \caption{MSMT17}
    \label{fig:distribution_camera_depression_angle_1}
  \end{subfigure}
  \hfill
  \begin{subfigure}{0.33\linewidth}
    \includegraphics[width=1.0\linewidth]{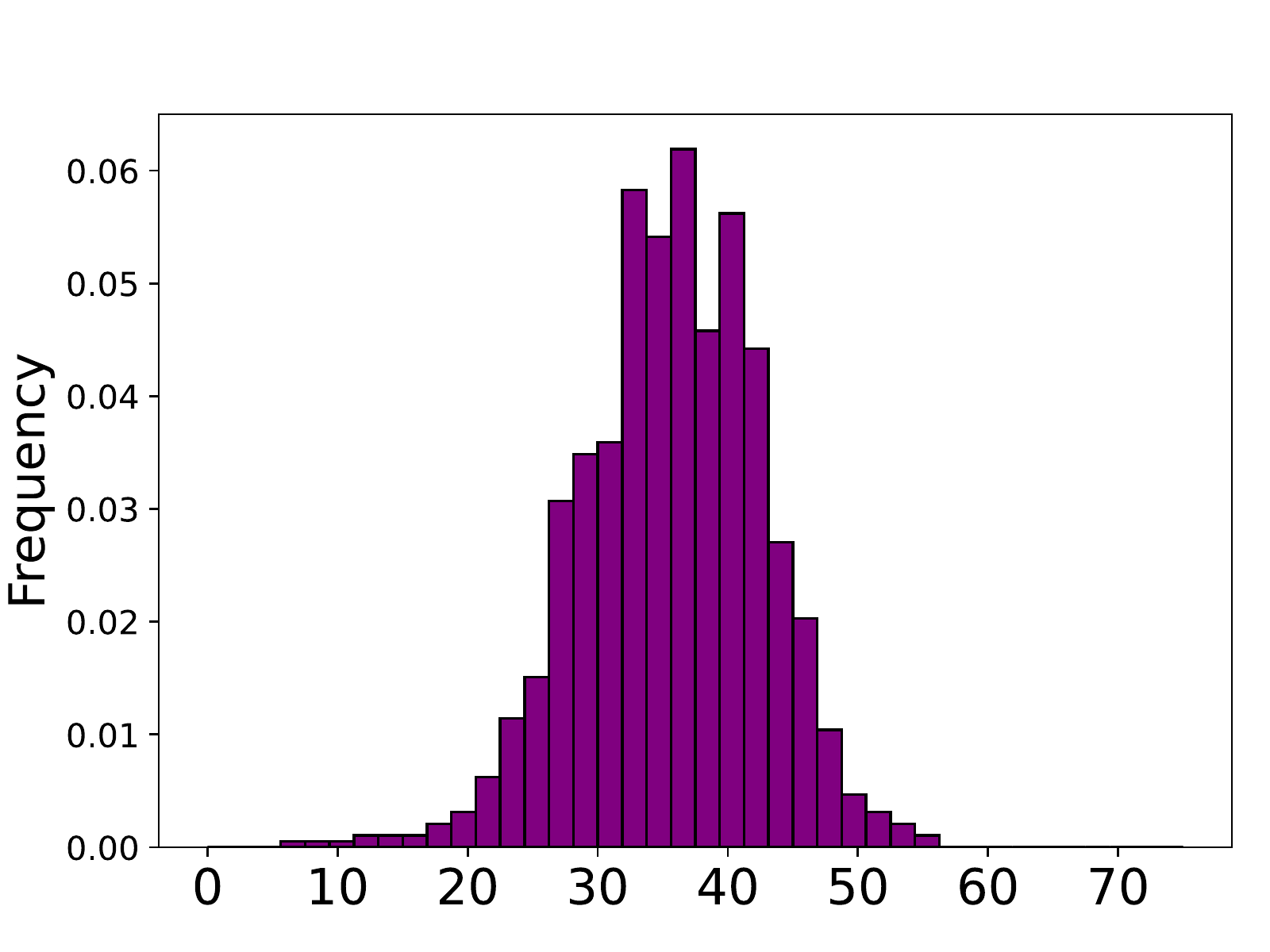}
    \caption{GRID}
    \label{fig:distribution_camera_depression_angle_1}
  \end{subfigure}
\vspace{-2mm}
\caption{Visualization of the distribution statistics of the camera depression angle in the Market-1510, MSMT17, and GRID datasets. From the statistical histogram we can see, the MSMT17 has a wider range of depression angles than Market-1501. The average value of the camera depression angles in the GRID dataset is larger.}
\vspace{-2mm}
\label{fig:distribution_camera_depression_angle}
\end{figure*}

\subsubsection{Camera Depression angle}
Few studies have looked at the effect of camera depression angle. We find that in the actual scene, there is a big difference in the heights of the installation position of the cameras, which has a great impact on the appearances of persons. For example, most of the camera shots of the MSMT dataset are in a flat direction, while the average value of camera depression angles in the GRID \cite{loy2009multi} dataset is larger. We train an estimation model for camera depression angle by the target-aware information extraction method. The model is used to estimate the depression angle value of each image in three datasets, and the distribution statistics are visualized in \cref{fig:distribution_camera_depression_angle}.

We compare the results of rendered TAGPerson with or without target domain information about the camera depression angle in \cref{table:da_depression_angle_and_gamma}. The column TAGP-C represents the rendering option value of camera depression angle is target-aware. We can find that compared to the basic setting if the distribution of the camera depression angle coincides with the specific domain, the performance of the trained ReID model can be improved. For example, the images of dataset GRID are captured underground, and the depression angle of the camera is obviously larger. The performance of TAGP-C rendered in a target-aware manner for GRID outperforms the basic method by $37.2\%$ vs. $28.8\%$ and $45.2\%$ vs. $38.2\%$ for rank-1 accuracy and mAP, respectively.

\begin{table}[t]
\setlength\tabcolsep{5.5pt}
\caption{Direct transfer performance on the Market, MSMT, and GRID datasets. The first row without checkmark symbols in TAGP-C and TAGP-G columns represents the manual setting for TAGPerson. TAGP-C represents that the rendering option about camera depression angle is guided by the target domain information. TAGP-G represents that the rendering option about gamma value is guided by the target domain information. The last row means that both the camera depression angle and gamma value are controlled by the target domain information.}
\vspace{-3mm}
\label{table:da_depression_angle_and_gamma}
\begin{center}
\footnotesize
\begin{tabular}{|c|c|cc|cc|cc|}
    \hline
    \multirow{2}{*}{TAGP-C} & \multirow{2}{*}{TAGP-G} & \multicolumn{2}{c|}{Market} & \multicolumn{2}{c|}{MSMT} & \multicolumn{2}{c|}{GRID}\\
    \cline{3-8} 
         & & R1 & mAP & R1 & mAP & R1 & mAP\\
    \hline
    \hline 
    & & 79.9 & 53.1 & 40.9 &14.3 & 28.8 &  38.2 \\
    \checkmark & & 81.2 & 54.5 & 46.2 & 17.2 & 37.2 & 45.2 \\
    & \checkmark & 81.3 & 54.0 & 46.3 & 17.1 & 38.0 & 45.9 \\
    \checkmark & \checkmark & \bfseries 81.6 & \bfseries 54.8 & \bfseries 47.5 & \bfseries 17.7 &  \bfseries 38.8 & \bfseries 47.3 \\
    \hline
\end{tabular}
\vspace{-3mm}
\end{center}
\end{table}

\subsubsection{Gamma Value}

Gamma correction is a nonlinear operation used to encode and decode luminance or tristimulus values in video or still image systems. For devices with different gamma correction settings, the captured images may appear in different brightness. 
The images captured by one camera may also show a wide range of brightness as the light changes from morning to evening. 
There are differences in gamma value parameters between different datasets because of the camera devices and acquisition time. 
For different distribution statistics, we construct the TAGPerson dataset in a target-aware manner towards the gamma value, which is constrained by information extracted from the target domain and controls the range of the rendering option values.

From \cref{table:da_depression_angle_and_gamma} we can also find that the performance of the ReID model can be improved by integrating the gamma information of the specific domains. The improvements to MSMT and GRID datasets are obvious. For the MSMT dataset, the rank-1 accuracy is improved from $40.9\%$ to $46.3\%$. For the GRID dataset, the target-aware setting can obtain $7.7\%$ mAP improvement compared to the manual setting. That makes sense because the images of the GRID dataset are captured underground and the illumination condition is poor. Meanwhile, many images in the MSMT dataset are taken in reverse light. Thus the acquisition of the gamma value from the target images can be useful.

\begin{figure*}
\centering
  \begin{subfigure}{0.48\linewidth}
    \includegraphics[width=1.0\linewidth]{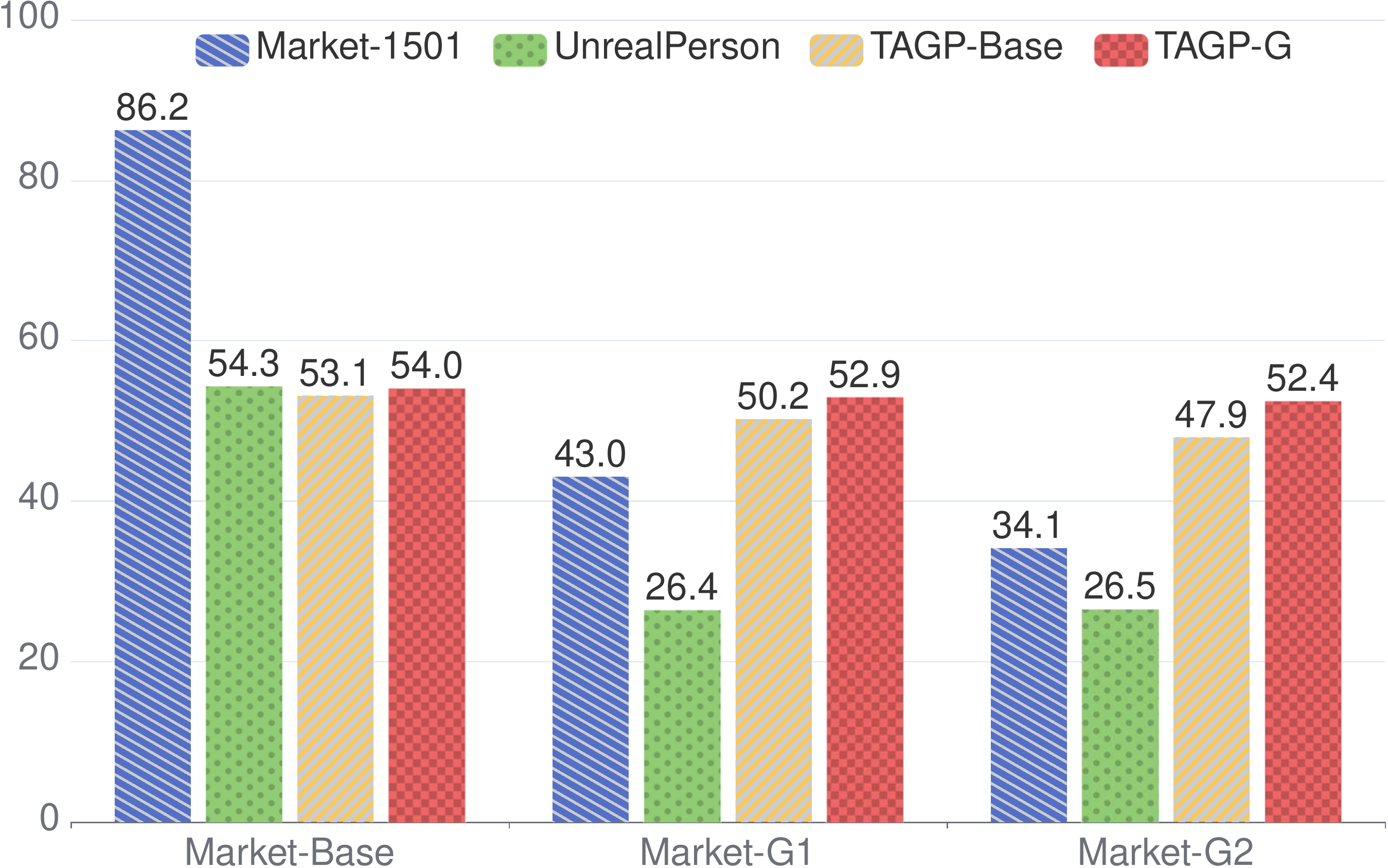}
    \caption{The mAP(\%) performance on Market-variant datasets.}
    \label{fig:da_gamma_market_and_msmt_1}
  \end{subfigure}
  \hfill
  \begin{subfigure}{0.48\linewidth}
    \includegraphics[width=1.0\linewidth]{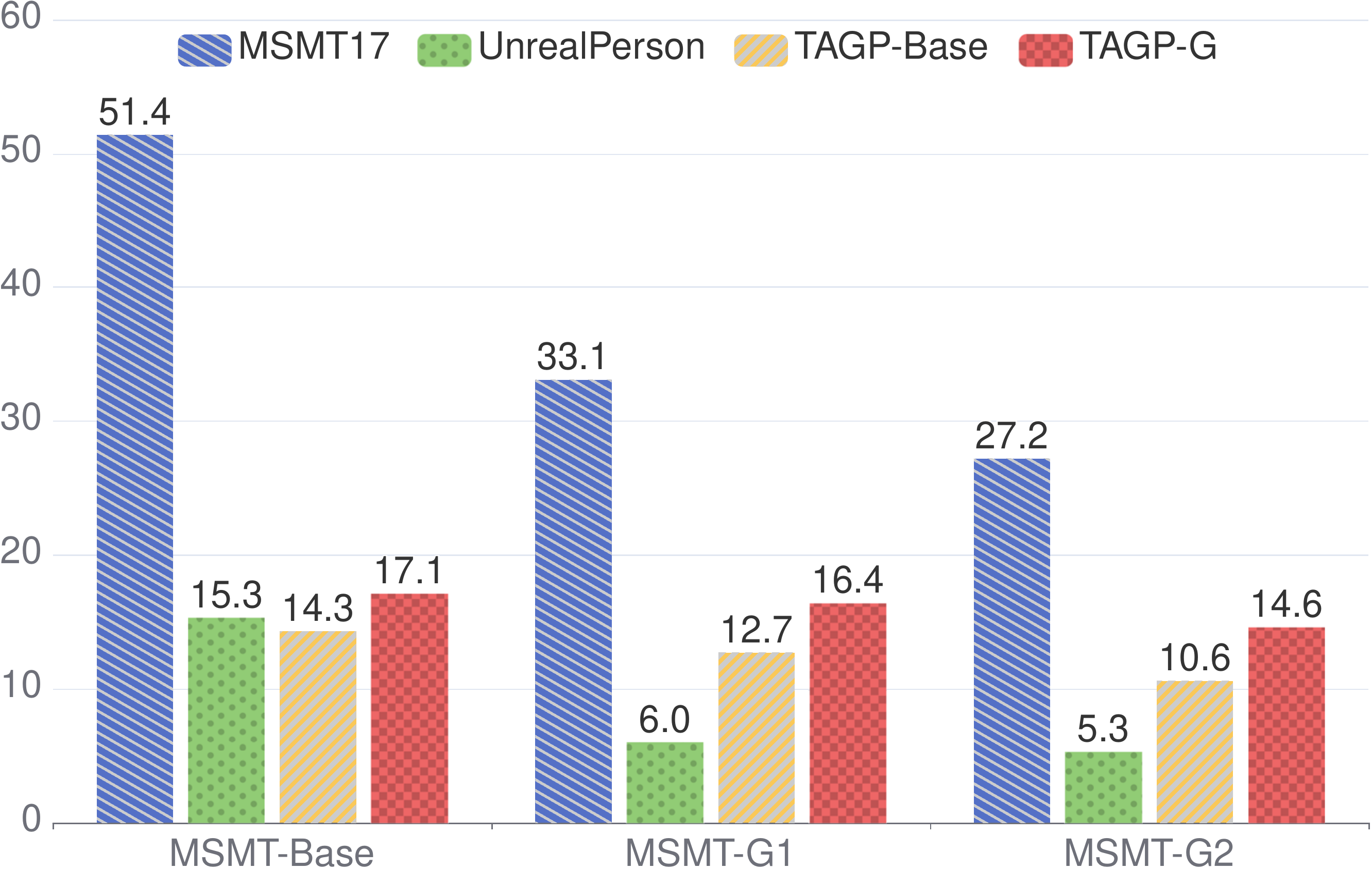}
    \caption{The mAP(\%) performance on MSMT-variant datasets}
    \label{fig:da_gamma_market_and_msmt_2}
  \end{subfigure}
\caption{The mAP performance of models which are trained from different training data is drawn in the bar chart. TAGP-Base represents images rendered in the manual setting and TAGP-G represents images rendered according to the gamma value distribution. (a) Test sets are Market variant datasets. (b) Test sets are MSMT variant datasets. Market-Base and MSMT-Base represent the original Market and MSMT datasets. When the gamma values are disturbed, the model trained from supervised learning and UnrealPerson datasets deteriorate significantly. TAGP-Base also drops slightly. The model trained from the TAGP-G dataset which is rendered in a target-aware manner keeps a steady performance on different target domains. This ensures the robustness of the model in practical application.}
\label{fig:da_gamma_market_and_msmt}
\end{figure*}

\subsubsection{Gamma Value for Extreme Scenarios}
Integrating gamma value has been proved to have a big boost for extreme scenarios like the GRID dataset. To explore the effects more deeply, we conducted extension experiments.
Due to the lack of similar large ReID datasets in extreme scenarios, we decide to simulate possible actual scenarios where the gamma values are changed. 
We create the variant datasets of gamma by applying different gamma values to the original images. For the Market dataset, we create Market-G1 datasets where the gamma values are randomized from $0.5$ to $1.5$, and Market-G2 datasets where the gamma values are randomized from $0.5$ to $2.0$. The same operation is performed for the MSMT dataset and we can get MSMT-G1 and MSMT-G2 variant datasets.

\cref{fig:da_gamma_market_and_msmt_1} shows the performance of different training datasets on these test sets. 
TAGP-Base represents the TAGPerson dataset rendered under manually set parameters. TAGP-G represents the TAGPerson dataset rendered in a target-aware manner about the gamma value.
We can see that, with only a little gamma jitter on images of the Market dataset, the mAP of the model trained from supervised learning significantly drops from $86.2\%$ to $43.0\%$. The performance of synthetic datasets like UnrealPerson or TAGP-Base also declines seriously. It seems that the gamma discrepancy introduces a large domain shift. The TAGP-G dataset is rendered in a target-aware manner by integrating the distribution statistics of the target domain, so it can alleviate the problem to some extent and achieve better performance in its corresponding case. Compared to the dataset generated with manually set rendering options, the target-aware one gains $4.5\%$ mAP improvement on the Market-G2 dataset. Note that the results of TAGP-G on Market-Base, Market-G1, and Market-G2 are not from the same model. There are three target domains and the TAGP-G serials are de facto three models on the target domain, respectively.

The same phenomenon can be observed for the MSMT dataset. The changes in gamma value significantly affect the performance of the ReID model. The results can be seen in \cref{fig:da_gamma_market_and_msmt_2}. By rendering images in a target-aware manner, the TAGP-G obtain $17.1\%$, $16.4\%$, and $14.6\%$ mAP on MSMT-Base, MSMT-G1, and MSMT-G2 datasets, respectively. Compared to using the manual setting, the TAGP-G reduces the model deterioration in extreme scenarios. We also conduct experiments to compare the effect between data augmentation and target-aware gamma, the conclusion is that they are complementary. We will analyze it in supplementary materials.

\subsection{Discussion and Limitation} 
One of the drawbacks of our TAGPerson is that there is a natural lack for situations of occlusion and multiple persons because the rendering process is person-centered. We have tried to deliberately add the occlusion items by using object annotations from COCO dataset \cite{lin2014microsoft}. 
Via using the instance segmentation annotations, the rendered images have a realistic occlusion effect. 
Besides, we have created the images containing multiple persons by placing another person nearby, and they can be rendered to be like partners. 
However, adding images of these two scenarios has not obtained improvement on the performance. Maybe we have not found the correct way to render this kind of situation.
We will discuss it in supplementary materials.

{\bf Broader Impact.} Person ReID technology may inevitably infringe the privacy of pedestrians. Our work attempts to reduce this infringement from two aspects. In the pre-training stage, the synthetic TAGPerson dataset can be used to replace real datasets, e.g., DukeMTMC-ReID, which has been taken down due to ethics issues. In terms of target domain information utilization, we use statistical information rather than raw data to avoid accessing images directly. However, the real images from surveillance data are necessary for the test stage when applying the model to actual scenarios. This may potentially raise privacy issues because not all human objects know and permit that they are being recorded. We urge that users should follow strict regulations and laws to use the person ReID models.

\section{Conclusion}
In this paper, we propose a target-aware generation pipeline named TAGPerson to resolve the person ReID task. Without establishing complex virtual scenes, we can directly render person images under the desired parameters, to serve as an effective training dataset. If the information of the target domain can be extracted, we can render the images in a target-aware manner by integrating the target domain information to guide the rendering options. This novel idea explores a new path to utilize the target domain when the images can not be accessed directly. TAGPerson provides a strategy to estimate environmental factors from the target images and an effective way to minimize the gap between synthesized datasets and real-world scenarios. In the future, we will study how to mine the potential key factors behind the rendering procedure.

{\small

\bibliographystyle{ieee_fullname}
}

\clearpage

\end{document}